\documentclass[conference]{IEEEtran}
\usepackage{cite}
\usepackage{amsmath,amssymb,amsfonts}
\usepackage{algorithmic}
\usepackage{graphicx}
\usepackage{textcomp}
\usepackage{xcolor}
\usepackage{array}
\usepackage{mdwmath}
\usepackage{mdwtab}
\usepackage{eqparbox}
\usepackage{url}
\usepackage[acronym]{glossaries}
\usepackage{longtable}
\usepackage{acronym} 
\usepackage{cite}
\usepackage{graphicx}
\usepackage[export]{adjustbox}
\usepackage{array}
\usepackage{tabu}
\usepackage{longtable}
\usepackage{booktabs}
\usepackage{array,multirow,makecell}
\usepackage[utf8]{inputenc}
\usepackage[T1]{fontenc}
\usepackage{caption}
\usepackage{subcaption}
\usepackage{babel,blindtext}
\usepackage{enumitem}
\usepackage{paralist}
\usepackage{amssymb}
\usepackage{pifont}
\usepackage{lscape}
\usepackage{afterpage}
\usepackage{graphicx}
\def\BibTeX{{\rm B\kern-.05em{\sc i\kern-.025em b}\kern-.08em
    T\kern-.1667em\lower.7ex\hbox{E}\kern-.125emX}}

\makeatletter
\newcommand{\linebreakand}{%
  \end{@IEEEauthorhalign}
  \hfill\mbox{}\par
  \mbox{}\hfill\begin{@IEEEauthorhalign}
}

\begin{document}

\title{

Deep Reinforcement Learning for Autonomous Vehicle Intersection Navigation
}
\DeclareRobustCommand*{\IEEEauthorrefmark}[1]{%
  \raisebox{0pt}[0pt][0pt]{\textsuperscript{\footnotesize\ensuremath{#1}}}}

\author{\IEEEauthorblockN{Badr Ben Elallid\IEEEauthorrefmark{1},
Hamza El Alaoui\IEEEauthorrefmark{2}, and
Nabil Benamar\IEEEauthorrefmark{1,2}}
\IEEEauthorblockA{\IEEEauthorrefmark{1}Moulay Ismail University, Meknes, Morocco.   
\IEEEauthorrefmark{2}Al Akhawayn University in Ifrane, Morocco.}
badr.benelallid@edu.umi.ac.ma, h.elalaoui@aui.ma, n.benamar@umi.ac.ma
}

%\author{\IEEEauthorblockN{Badr Ben Elallid}
%\IEEEauthorblockA{\textit{School of Technology,} \\
%\textit{ Moulay Ismail University of Meknes}\\
%Meknes, Morocco \\
%badr.benelallid@edu.umi.ac.ma}

%\and

%\IEEEauthorblockN{Hamza El Alaoui}
%\IEEEauthorblockA{\textit{School of Science and Engineering,} \\
%\textit{Al Akhawayn University in Ifrane}\\
%Ifrane, Morocco \\
%h.elalaoui@aui.ma}

%\and

%\IEEEauthorblockN{Nabil Benamar}
%\IEEEauthorblockA{\textit{Moulay Ismail University of Meknes}\\
%\textit {School of Science and Engineering,} \\
%\textit{Al Akhawayn University in Ifrane}\\
%Ifrane, Morocco \\
%n.benamar@aui.ma}

%\linebreakand
 
%}

\markboth{Journal of \LaTeX\ Class Files,~Vol.~14, No.~8, August~2021}%
{Shell \MakeLowercase{\textit{et al.}}: A Sample Article Using IEEEtran.cls for IEEE Journals}

\IEEEpubid{0000--0000/00\$00.00~\copyright~2021 IEEE}

\maketitle

\begin{abstract}
In this paper, we explore the challenges associated with navigating complex T-intersections in dense traffic scenarios for autonomous vehicles (AVs). Reinforcement learning algorithms have emerged as a promising approach to address these challenges by enabling AVs to make safe and efficient decisions in real-time. Here, we address the problem of efficiently and safely navigating T-intersections using a lower-cost, single-agent approach based on the Twin Delayed Deep Deterministic Policy Gradient (TD3) reinforcement learning algorithm. We show that our TD3-based method, when trained and tested in the CARLA simulation platform, demonstrates stable convergence and improved safety performance in various traffic densities. Our results reveal that the proposed approach enables the AV to effectively navigate T-intersections, outperforming previous methods in terms of travel delays, collision minimization, and overall cost. This study contributes to the growing body of knowledge on reinforcement learning applications in autonomous driving and highlights the potential of single-agent, cost-effective methods for addressing more complex driving scenarios and advancing reinforcement learning algorithms in the future.

\end{abstract}

\section{Introduction}
 
Intersections present a considerable challenge to road safety due to their intricate traffic conditions, accounting for 36\% of road collisions \cite{milanes2009controller}. Conventional methods of controlling vehicle flow, such as traffic lights and stop signs, often hinder traffic progression and restrict intersection capacity. Autonomous driving strategies hold the potential to enhance intersection navigation by minimizing collisions caused by human error and optimizing driving behavior to reduce travel times \cite{huang2006low,elallid2022comprehensive}.

Existing intersection navigation solutions in autonomous driving predominantly depend on vehicle-to-vehicle (V2V) communication and centralized systems. In this paper, we investigate the prospects of a single-agent approach employing Reinforcement Learning (RL) techniques to develop advanced, adaptive, and cost-efficient solutions for traversing complex intersections \cite{elallid2022deep, elallid2022dqn}.

In this study, we apply the Twin-Delayed DDPG algorithm (TD3) to facilitate autonomous vehicles' (AVs) intersection navigation without relying on V2V communication or centralized systems. Our model trains AVs to arrive at their destinations without collisions by processing features extracted from images produced by the vehicle's front camera sensor and employing TD3 to predict the optimal action for each state. We simulate and train the proposed method using the CARLA simulator. Simulations results demonstrate the capacity of our model to learn over episodes in terms of reducing travel delay and collision rate.

The remainder of the paper is structured as follows: Section II reviews related works; Section III presents the problem formulation, including the state space, actions, and reward functions; Section IV details the simulation setup and experimental design; Section V presents our simulation results. Finally, Section VI concludes the paper and highlights future work and potential research directions.
  
\section{Related Works}
In recent years, autonomous driving research has advanced significantly, particularly in the area of intersection management. Two primary solutions have emerged: vehicle-to-vehicle (V2V) communication and centralized solutions. Concurrently, reinforcement learning (RL) algorithms have gained traction within autonomous driving, especially for navigation tasks. However, their application in intersection navigation remains under investigated. Thus, while V2V and centralized strategies dominate the literature, RL-based approaches present a promising yet unexplored research direction \cite{elallid2023vehicles}.
\subsection{Vehicle-to-Vehicle Solutions}
V2V communication has been proposed as a way to improve traffic throughput and safety at intersections for autonomous vehicles. In\cite{azimi2013reliable}, the authors investigate V2V communication for cooperative driving, particularly in intersection management. They propose V2V intersection protocols that not only enhance traffic throughput but also prevent deadlock situations. Simulation results demonstrate considerable improvements in both performance and safety.

Another study \cite{li2006cooperative} explores cooperative driving at blind intersections, which lack traffic lights, utilizing intervehicle communication. The authors introduce safety driving patterns representing collision-free movements and develop trajectory planning algorithms aiming to minimize execution time. Simulated results underscore the potential and utility of the proposed algorithms.

\subsection{Centralized Solutions}
Centralized approaches to intersection management typically rely on a Roadside Unit (RSU) to coordinate the crossing sequence. However, this can be costly, as each intersection would require an RSU. For example, in \cite{riegger2016centralized}, the authors present a centralized Model Predictive Control (MPC) approach for the optimal control of autonomous vehicles within an intersection control area. The problem is formulated as a convex quadratic program, enabling efficient solutions.

In \cite{chen2019autonomous}, researchers propose an autonomous T-intersection strategy that combines motion-planning and path-following control, considering oncoming vehicles. Through CarSim simulations \cite{carsim} and scaled car experiments, the effectiveness of the motion planner in generating collision-free trajectories and the path-following controller in ensuring safe and swift intersection navigation is demonstrated.

In \cite{medina2015automation}, a Cooperative Intersection Control (CIC) methodology is developed to improve T-intersection navigation for autonomous vehicles. By employing virtual platoons and simulating six vehicles crossing the intersection, the authors demonstrate increased traffic efficiency without stopping.

Although considerable progress has been made in intersection management through V2V communication and centralized solutions, exploring alternative approaches remains crucial. Employing reinforcement learning (RL) techniques within a single-agent framework for intersection navigation has the potential to produce adaptive and advanced solutions. These solutions could not only improve the efficiency and safety of autonomous vehicles in intricate intersections but also potentially offer cost advantages over multi-agent and centralized methods.

\subsection{Reinforcement Learning Algorithms}

Reinforcement learning (RL) is a branch of machine learning focused on training agents to make decisions by interacting with their environment. In RL, an agent learns an optimal policy, which maps states to actions, by maximizing the cumulative reward it receives from the environment. This learning process typically involves exploring the environment to gather information and exploiting the knowledge acquired to optimize actions \cite{sutton2018reinforcement}

Popular RL algorithms, such as Deep Q-Network \cite{mnih2015human,elallid2023reinforcement}, Deep Deterministic Policy Gradient \cite{lillicrap2015continuous}, Proximal Policy Optimization \cite{schulman2017proximal}, and Soft Actor-Critic \cite{haarnoja2018soft}, have been employed to address various challenges in autonomous driving, including intersection navigation. However, their performance and suitability for different driving scenarios can vary significantly.

Twin Delayed DDPG \cite{fujimoto2018addressing} has emerged as a relevant and promising algorithm for autonomous vehicles due to its stability, reduced overestimation bias, and improved exploration capabilities. TD3 is an off-policy actor-critic algorithm that extends DDPG by incorporating three key enhancements: 
\subsubsection{Twin Q-networks}which are used to mitigate overestimation bias by maintaining two separate Q-function approximators and taking the minimum value of the two.
\subsubsection{Delayed policy updates}wherein the actor and target networks are updated less frequently than the Q-networks to improve stability.
\subsubsection{Target policy smoothing}which adds noise to the target actions during the learning process to encourage exploration and prevent overfitting to deterministic policies.

These improvements have enabled TD3 to achieve superior performance in a variety of tasks, including intersection management, compared to its predecessor DDPG \cite{fujimoto2018addressing}. Additionally, TD3 offers an efficient approach for learning complex decision-making policies required for navigating intersections safely, making it well-suited for autonomous driving applications.

\subsection{Simulation Environments for Autonomous Driving Research}

Various simulation platforms have been developed to facilitate the evaluation of reinforcement learning (RL) algorithms for autonomous driving applications. Widely used platforms include CARLA, SUMO, Gazebo, CarSim, and LGSVL. Each platform offers unique advantages and capabilities tailored to different aspects of autonomous driving research.

CARLA (Car Learning to Act) is an open-source simulator specifically designed for autonomous driving research, providing high-fidelity urban environments, diverse traffic scenarios, and a range of weather conditions. It enables comprehensive testing and validation of autonomous driving algorithms, including RL approaches \cite{Dosovitskiy17}.

SUMO (Simulation of Urban Mobility) is a microscopic traffic simulator that models individual vehicles and their behavior in various traffic situations. It is particularly beneficial for large-scale simulations and can be integrated with other platforms or tools, allowing researchers to investigate the impact of RL algorithms on overall traffic flow and intersection management \cite{krajzewicz2012recent}.

Gazebo is a versatile and extensible 3D robotics simulator that supports multiple physics engines, sensor models, and control interfaces. It facilitates the simulation of intricate autonomous driving scenarios, making it well-suited for the evaluation of RL algorithms in dynamic and challenging environments \cite{koenig2004design}.

CarSim is a commercial vehicle dynamics simulation software that offers high-fidelity vehicle models and realistic driving environments. It allows for the integration of control systems and sensors, enabling researchers to evaluate the performance of RL algorithms in terms of vehicle control and dynamics \cite{carsim}.

LGSVL is a multi-robot AV simulator developed by LG Electronics America R\&D Center, providing an out-of-the-box solution for testing autonomous vehicle algorithms. It offers photo-realistic virtual environments, sensor simulation, and vehicle dynamics, and supports integration with popular AD stacks such as Autoware and Baidu Apollo \cite{LGSVL}.
 
In our research, we opted for the CARLA simulator platform, given its focus on autonomous driving and comprehensive feature set. CARLA offers high-fidelity urban environments, diverse traffic scenarios, and a range of weather conditions, enabling rigorous testing and validation of our reinforcement learning (RL) algorithms\cite{Dosovitskiy17}. Additionally, CARLA's open-source nature and active community support make it a highly accessible and extensible tool for our research purposes. The platform's compatibility with Python allows seamless integration with machine learning frameworks, such as PyTorch and TensorFlow, streamlining the development process. Consequently, the combination of CARLA and Python provides a powerful and accessible foundation for our autonomous driving research.
 
%======================= Figure of proposed method ====================
\begin{figure*}[htp]
    \centering
   \includegraphics[width=\textwidth,height=9cm,keepaspectratio]{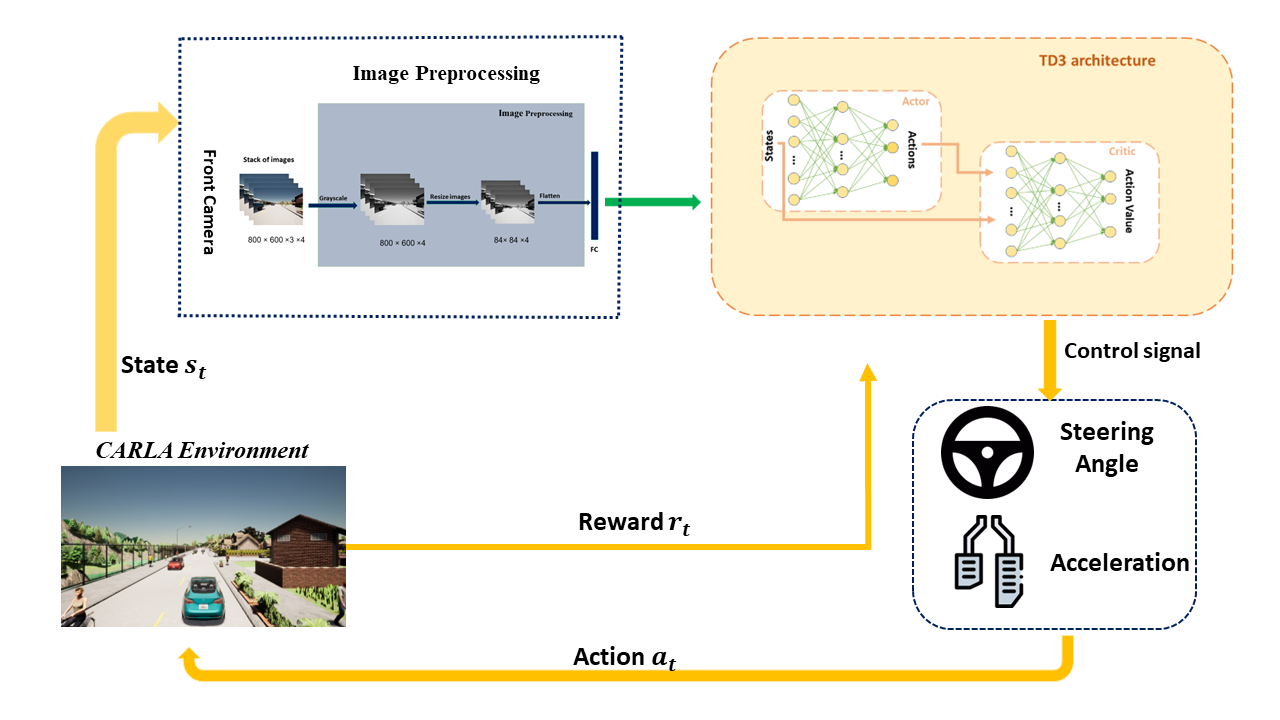}
    \caption{TD3-based Deep Reinforcement Learning architecture for controlling AV in a T-intersection scenario with dense traffic.} 
    \label{fig:archi}
\end{figure*} 

\section{Problem Formulation}
\begin{compactitem}[$\bullet$]

     \item \textbf{State space}: 
      In the real world, human drivers rely on more than just their visual perception to comprehend their surroundings; they also take into account the motion of other road users. Analogously, an autonomous vehicle (AV) must utilize a sequence of images to grasp the movement of objects within its environment. In this particular instance, our model processes a series of four consecutive RGB images acquired by the AV's front camera. These images have dimensions of $800 \times 600 \times 3 \times 4$ pixels, which we subsequently resize to $84 \times 84 \times 3 \times 4$ pixels and convert into grayscale. The resulting state $S_{t}$ possesses dimensions of $84 \times 84 \times 4$, which are then input into our Actor and Critic architecture.
      \item \textbf{Action space}: 
       The AV in the CARLA simulator receives three control commands from the environment: acceleration, steering, and braking. These commands are represented by float values ranging $[0,1]$ for acceleration, $[-1,1]$ for steering, and $[0,1]$ for braking. Since TD3 is a continuous DRL algorithm, the agent must select actions continuously. Therefore, at each step $t$, the agent needs to choose an action represented by (acceleration, steering, brake) while ensuring that each command falls within its respective range.
       \item \textbf{Reward function} :
        In our scenario, we formulate a reward function that takes into account potential situations in urban traffic. The AV must not collide with other participants on the road, such as vehicles, cyclists, motorcycles, and pedestrians, while also successfully reaching its intended destination. For this purpose, the reward function includes four components: a reward for maintaining speed, a penalty for collisions,  driving in other or off the road, and the distance to the goal. We can represent the current distance to the goal as $D_{cu}$, the previous distance to the goal as $D_{pre}$, the velocity of the AV as $V_{speed}$, and the measure of the AV being off or in another lane as $Me_{offroad}$ and $M_{otherlane}$, respectively. 
        \begin{equation*}
         reward = \left\{
         \begin{array}{llll}
             R_{t1}  = - C_{collison} \\
             R_{t2}  = D_{pre} - D_{cu} \\
             R_{t3}  =  \max(0, \min(V_{speed}, V_{limit})) \\
             R_{t4}  =  -M_{offroad} - M_{otherlane} \\
             R_{t5}  = 100 \\
             R_{t}   =  R_{t1} +  R_{t2} + R_{t4} + R_{t5}
            \end{array}
            \right.
            \end{equation*}
        Where $V_{limit}$ speed limit and  $C_{collison}$ is the penalty for colliding with road users such as vehicles, pedestrians, cyclists, and motorcycles.  
        \item \textbf{Training}: Our TD3-based deep reinforcement learning architecture comprises of two networks: actor and critic. The actor network takes the current state as input and generates the next action for the agent. On the other hand, the critic network predicts the action value based on both the state and the value obtained from the actor. To encourage exploration during training, the actor network adds noise to the predicted action. We have set the number of hidden layers to two, with each layer containing 256 neurons. Other parameters are presented in Table \ref{tab1}
\end{compactitem}
\vspace{\baselineskip}

 %================ Figure of Our Scenario =====================
\begin{figure*}%
    \centering
    \subfloat[\centering  ]{{\includegraphics[width=10cm,keepaspectratio]{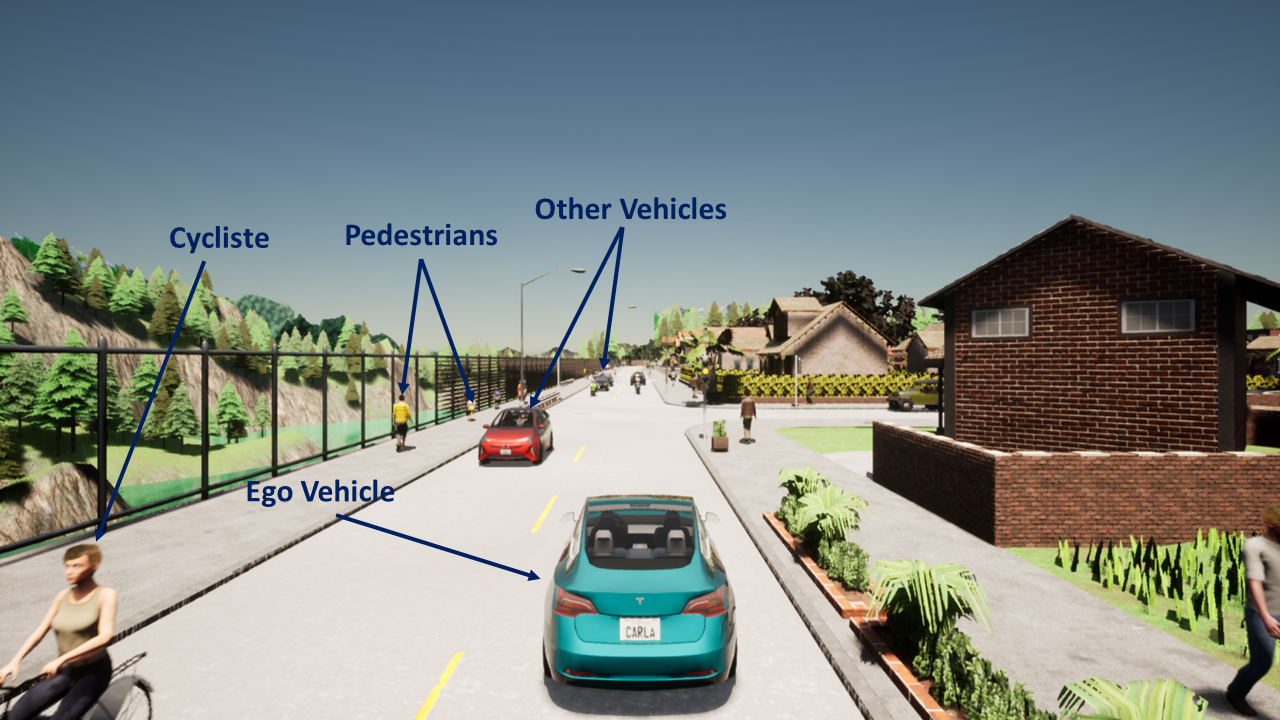} }}%
    \qquad
    \subfloat[\centering  ]{{\includegraphics[width=10cm,keepaspectratio]{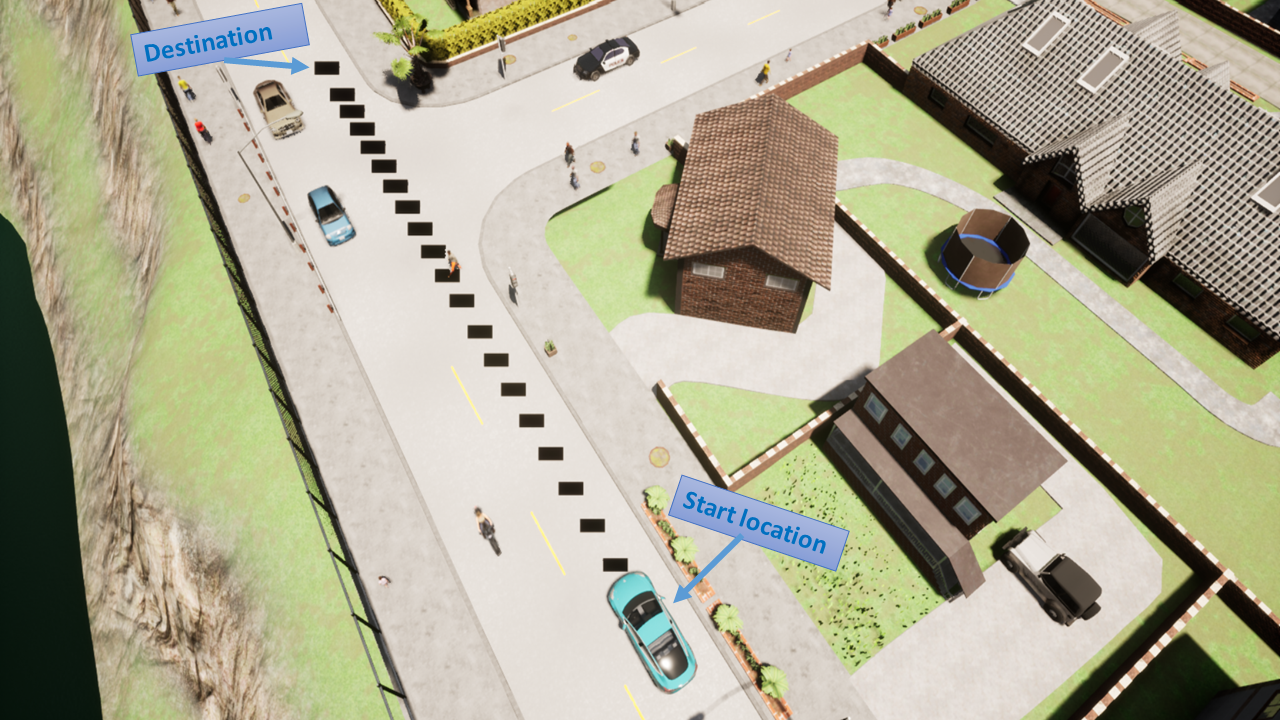} }}%
    \caption{The scenario we considered : (a) The ego vehicle navigating through an T-intersection with heavy traffic; (b) The path that the autonomous vehicle (AV) is supposed to follow to reach its desired destination.}%
    \label{fig:scenario}%
\end{figure*}

 \section{Simulation Setup}
 
We employed the CARLA simulator (version 0.9.10) and PyTorch for our autonomous driving experiments, focusing on a T-intersection scenario within the CARLA environment. This scenario presents a complex and challenging situation for autonomous driving systems due to its unique traffic dynamics.

Vehicles in the T-intersection scenario must navigate multiple lanes while interacting with other road users, such as managing oncoming traffic, merging with traffic flow, and responding to pedestrians at designated crossing locations. The numerous real-time decisions and actions required for safe and efficient navigation contribute to the scenario's complexity.

To create a comprehensive and realistic simulation, we included various traffic participants like pedestrians, cars, bicycles, and motorcycles. This diverse set of road users enables a thorough assessment of the autonomous driving algorithm's performance across a wide range of traffic situations and challenges.

We adjusted parameters such as vehicle speeds, distances between vehicles, and the frequency of traffic participants entering the simulation to emulate dense traffic conditions. This allowed us to subject our algorithms to demanding circumstances that closely resemble real-world driving conditions, essential for evaluating the effectiveness of our proposed solutions in complex driving environments.

Figure \ref{fig:scenario}, illustrates a dense traffic environment where in an autonomous vehicle navigates safely through a T-intersection. The vehicle starts at an initial position, follows a designated path to its destination, and avoids collisions with pedestrians, cyclists, motorcycles, and other vehicles. The scenario involves 300 randomly moving vehicles, including motorcycles, cyclists, and four-wheeled vehicles, managed by the CARLA traffic manager. All vehicles are set to autopilot mode with a safe distance of 2.5 meters between them and speed limit equal 30 m/s. We also randomly place 100 pedestrians, with 80\% of them crossing the road using a crosswalk, adding to the challenge. Training episodes terminate under the following conditions: 1) collision between the AV and other road users; 2) the AV successfully reaches its destination; 3) the episode exceeds the maximum number of training steps, which is set to 500.

\begin{table}[htp]
 \centering
\caption{Parameters used in the simulation}
\label{tab1}
\begin{tabular}{ p{4cm}p{3cm}}
\hline
\textbf{Parameter}& \textbf{Value} \\ \hline
Learning rate (actor \& critic) & 0.0003 \\ \hline
Episodes & 2000 \\ \hline
Batch size & 64 \\ \hline
$\gamma$ &  0.99 \\  \hline
Exploration noise &  0.1 \\  \hline
Exploration step &  10000 \\  \hline
Policy update frequency &  2 \\  \hline
Replay Memory Size & 5000 \\ \hline
\end{tabular}
\end{table}

\begin{figure}[htp]
  \includegraphics[width=\linewidth,keepaspectratio]{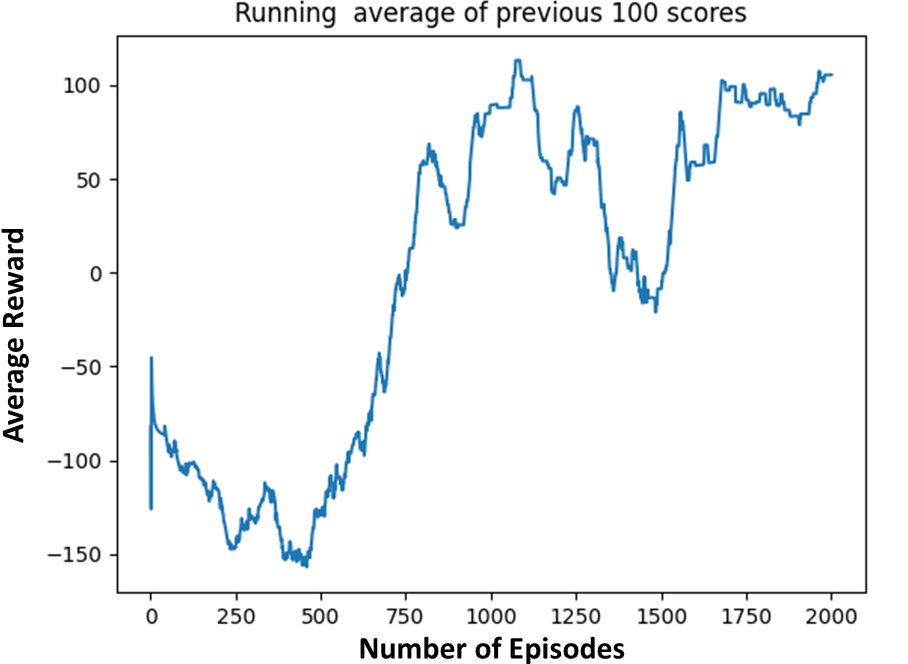}
  \caption{Average reward during episodes}\label{fig:av_reward}
\end{figure}

\section{Results \& Discussion}
In this section, we present the results obtained using our proposed method. Figure \ref{fig:av_reward} illustrates that the average reward progressively increases over episodes, ultimately attaining a high value at 2000 episodes. Furthermore, around the 2000th episode, the model stabilizes and converges, showcasing the efficacy of our approach.

During the testing phase, the T-intersection remains consistent with the training phase, though the traffic density varies. The density is determined by the random spawning of pedestrians (Ped) and other vehicles, such as cyclists and motorcycles (Veh), within the environment. We select five scenarios: 1) $Ped = 100$ and $Veh = 100$; 2) $Ped = 200$ and $Veh = 200$; 3) $Ped = 300$ and $Veh = 300$; 4) $Ped = 400$ and $Veh = 400$; 5) $Ped = 450$ and $Veh = 450$.

\begin{figure}[htp]
  \includegraphics[width=\linewidth,keepaspectratio]{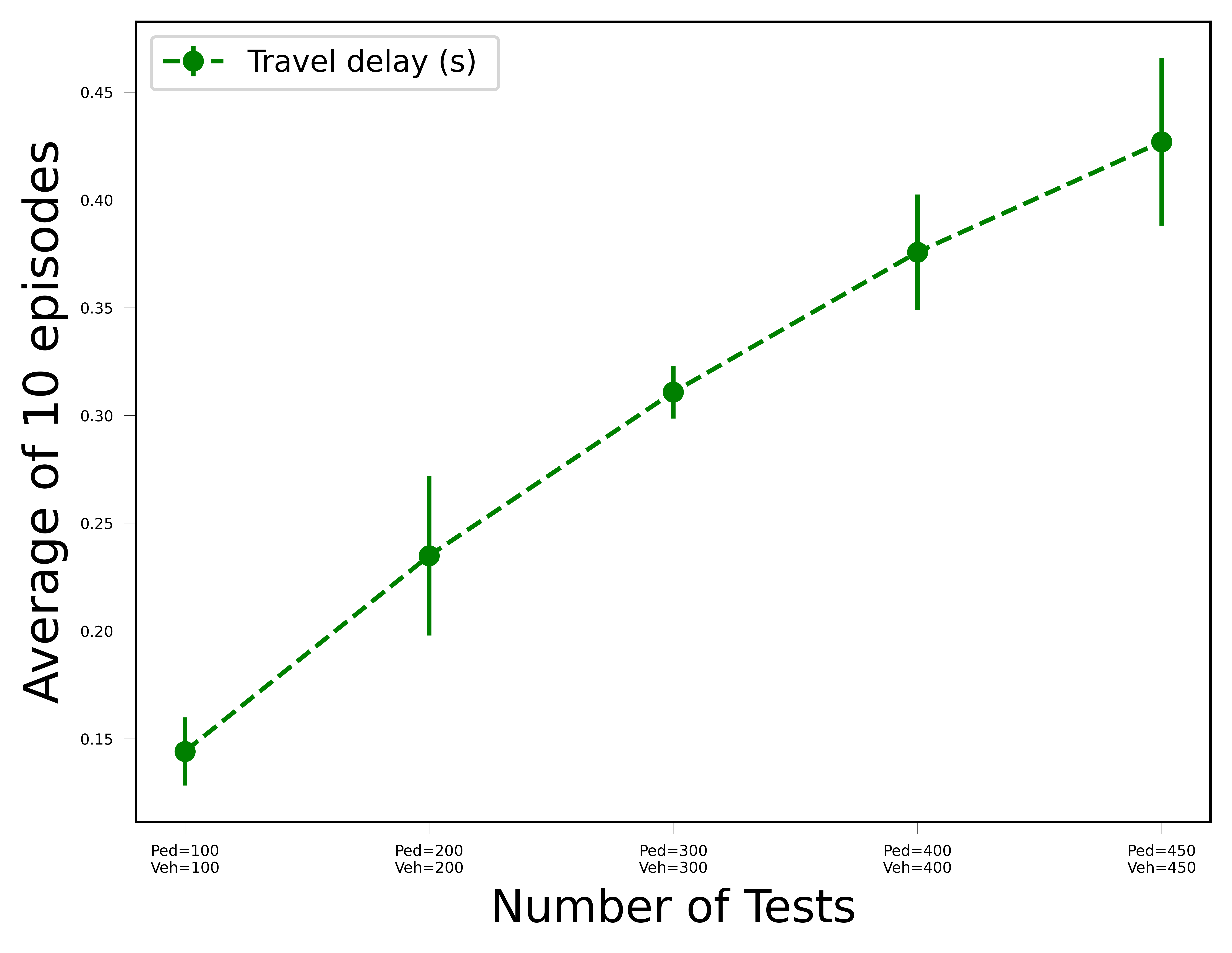}
  \caption{The average of travel delay each test}\label{fig:test_Travel}
\end{figure}

\begin{figure}[htp]
  \includegraphics[width=\linewidth,keepaspectratio]{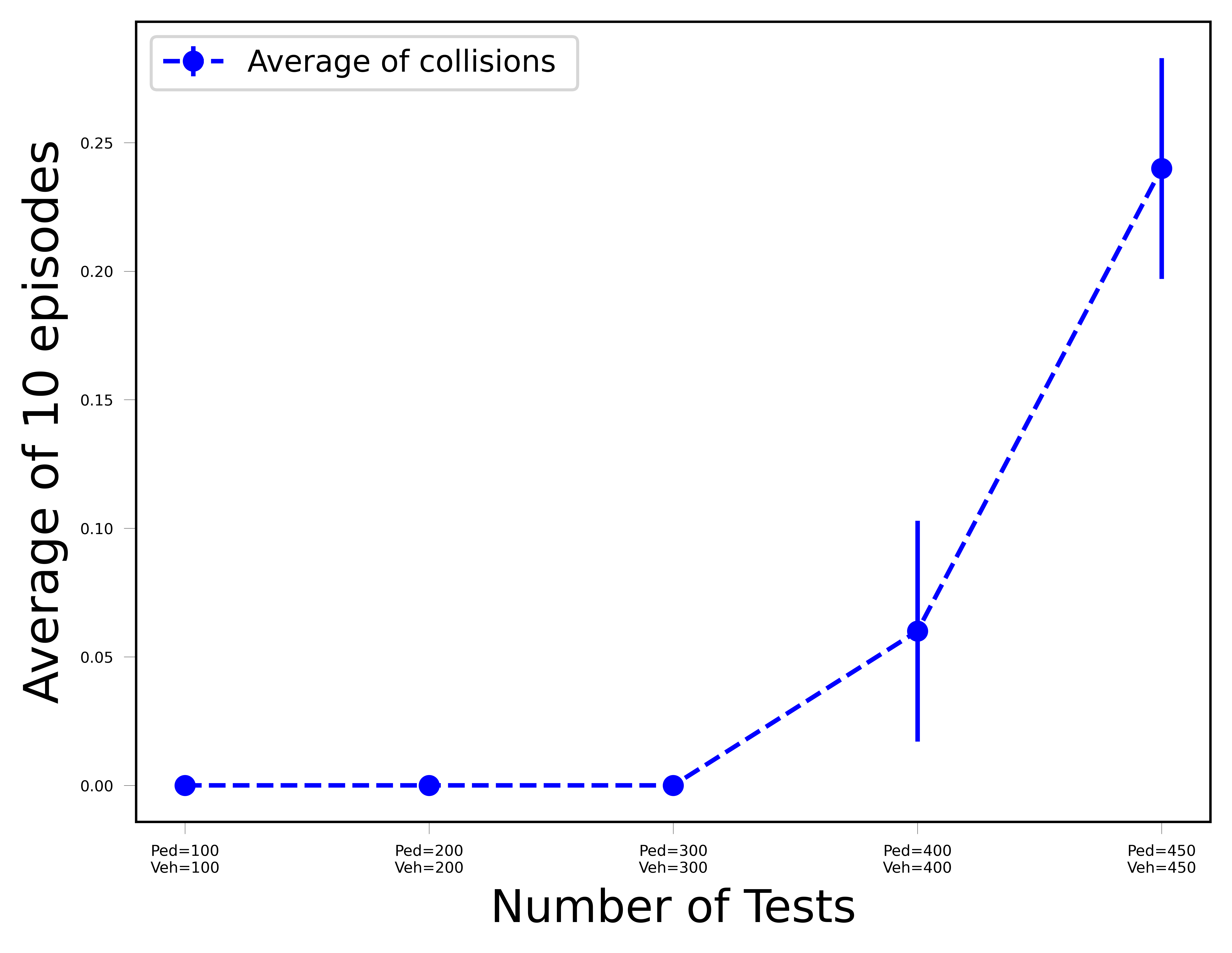}
  \caption{The average of collisions in each test}\label{fig:test_Crashe}
\end{figure}

The policy network trained by the model governs the vehicle as it navigates its environment and reaches its destination during the testing phase. To evaluate the model, we execute ten episodes and measure the travel delay and number of collisions in each episode. Since the model's objective is to minimize travel delay and accidents in dense traffic, we compute the average of both metrics. We repeat this process ten times and establish the confidence interval. The test results, depicted in Figures \ref{fig:test_Travel} and \ref{fig:test_Crashe}, demonstrate that the vehicle quickly arrives at its destination and avoids collisions. Our model exhibits a distinct advantage in evading collisions with road participants, as the average number of collisions remains low.

\section{Conclusion}
In this paper, we presented single-agent approach for navigating complex T-intersections using Twin Delayed Deep Deterministic Policy Gradient (TD3) in a dense traffic scenario. Our proposed method employs reinforcement learning to train an autonomous vehicle (AV) to make safe and efficient decisions in real-time. We leveraged the CARLA simulation platform to create a realistic urban environment, featuring diverse traffic participants and challenging driving conditions.

Our results indicate that the TD3 algorithm demonstrates stable convergence and improved exploration capabilities, enabling the AV to navigate T-intersections safely and effectively. Furthermore, the proposed method exhibits a low number of collisions and reduced travel delays in various traffic density scenarios, highlighting its potential for real-world autonomous driving applications.

In future work, we aim to explore the integration of additional sensors, such as LIDAR and RADAR, to enhance the AV's perception capabilities. Moreover, we plan to extend our research to more complex driving scenarios and investigate other advanced reinforcement learning algorithms for improved performance and robustness. Additionally, we intend to explore the impact of different network architectures and hyperparameters on the performance of the proposed method.

\bibliographystyle{ieeetr}
\bibliography{refs}

\vfill

\end{document}